\documentclass{article}
\usepackage[preprint]{neurips_2026}  


\usepackage[utf8]{inputenc}  
\usepackage[T1]{fontenc}  
\usepackage{hyperref}
\usepackage{url}
\usepackage{booktabs}
\usepackage{amsfonts}
\usepackage{amsmath}
\usepackage{nicefrac}
\usepackage{microtype}
\usepackage{xcolor}
\usepackage{subcaption}
\usepackage{graphicx}
\graphicspath{{./}}

\usepackage{fancyhdr}
\fancypagestyle{fermilabpub}{%
  \fancyhf{}%
  \fancyhead[R]{FERMILAB-PUB-26-0717-CMS-LDRD}%
  \fancyfoot[C]{\thepage}%
}

\newcommand{\partmodel}{ParT}
\newcommand{\moe}{MoEParT}

\title{Conditional Capacity and Routing in Mixture-of-Experts Particle Transformers}
\author{
\mdseries
Kaushik Pendiyala\textsuperscript{1,*} \quad
Haris Zia\textsuperscript{2,*} \quad
Trevin Lee\textsuperscript{2,*} \quad \\
Timothy Legge\textsuperscript{2} \quad
Alejandro J. De Leon \textsuperscript{2} \quad
Zihan Zhao\textsuperscript{2} \quad
Aaron Wang\textsuperscript{3} \quad \\ 
Abhijith Gandrakota\textsuperscript{4} \quad
Jennifer Ngadiuba\textsuperscript{4} \quad
Richard Cavanaugh\textsuperscript{4} \quad
Javier Duarte\textsuperscript{2} \quad \\ 
\textsuperscript{1}University of California Davis, Davis, CA, USA\\
\textsuperscript{2}University of California San Diego, San Diego, CA, USA\\
\textsuperscript{3}University of Illinois at Chicago, Chicago, IL, USA\\
\textsuperscript{4}Fermi National Accelerator Laboratory, Batavia, IL, USA\\ 
\textsuperscript{*}These authors contributed equally to this work.
}

\begin{document}
\maketitle
\thispagestyle{fermilabpub}
\begin{abstract}
Mixture-of-Experts (MoE) models can increase parameter capacity without proportionally increasing active computation, but it is unclear how this trade-off behaves in particle-physics transformers. We study dense and MoE Particle Transformers on 188-class JetClass-II, varying expert count, routing capacity, top-$K$, and auxiliary loss. We find that, when token dropping is avoided, top-1 MoE models improve over the dense baseline at nearly unchanged nominal forward compute, while further increasing the number of stored experts produces little additional accuracy gain. Activating multiple experts per token yields additional predictive improvements at higher computational cost. Routing analyses show that expert assignments become more strongly associated with particle identity and kinematics in some configurations, but this structure does not increase monotonically with classification performance. These results highlight the need to distinguish stored parameter capacity, active computation, routing capacity, and routing organization when evaluating sparse expert models for jet classification. Code and experiment configurations are available at \url{https://github.com/kpendiyala/MPT}.
\end{abstract}

\section{Introduction}
Particle Transformer (ParT) incorporates pairwise particle information into attention for jet classification \citep{qu2022}. Extending classification to the diverse resonance signatures in JetClass-II \citep{li2024} raises the question of how to allocate model capacity. Dense scaling increases parameters and computation together. Sparse mixture-of-experts (MoE) layers select a few expert networks per token, allowing parameter growth without proportional growth in active computation \citep{shazeer2017,fedus2022}. Routed language-model scaling studies explicitly treat parameter count and computation as separate axes \citep{clark2022}; we examine this distinction in constituent-level jet classification.

However, this does not guarantee better predictions. Restricted expert capacity can cause token assignments to be dropped, while more active experts change both computation and expert output combinations. Simultaneously, routing according to identifiable particle properties does not necessarily improve whole-jet classification. We therefore evaluate predictive performance together with routing behavior to determine whether more structured routing accompanies improved classification.

We compare dense \partmodel\ and \moe\ across expert count, capacity, top-$K$, and auxiliary loss. We find that no-drop top-1 expert scaling changes routing without improving accuracy, whereas top-2 improves prediction at higher nominal compute. Capacity constraints accompany degraded performance, and stronger routing association is not a monotonic indicator of better classification. Together, these results separate the effects of stored capacity, active computation, and routing structure in MoE-based jet classifiers.

\section{Related Work}
\paragraph{Deep learning for jet tagging.}
Modern jet-tagging methods increasingly operate directly on sets of reconstructed particles rather than relying only on engineered jet observables. ParticleNet represents jets as particle clouds and applies dynamic graph convolution, providing a strong benchmark for constituent-based jet classification \citep{qu2019particlenet}. Particle Transformer (ParT) subsequently introduced a Transformer architecture that incorporates pairwise particle interactions into the attention mechanism and achieved improved performance on the large-scale JetClass benchmark \citep{qu2022}. Our work retains the ParT representation and attention structure while modifying the feed-forward sublayers through sparse mixture-of-experts routing.

\paragraph{Large-scale jet pre-training and JetClass-II.}
JetClass-II extends large-scale jet classification to a substantially broader set of boosted-jet signatures. The Sophon framework uses a Particle Transformer pre-trained on a 188-class JetClass-II classification task and demonstrates the utility of the learned representation for heavy-resonance searches and transfer learning \citep{li2024}. In contrast, our study uses the JetClass-II classification task as a controlled setting for comparing dense and sparse-expert Particle Transformers, focusing on conditional parameter capacity, routing capacity, active computation, and routing organization rather than on downstream transfer.

\paragraph{Sparse mixture-of-experts models.}
Mixture-of-experts models use learned routing to activate only a subset of model parameters for each input, increasing stored parameter capacity without requiring all parameters to be evaluated for every token. Early sparsely gated MoE models demonstrated large increases in parameter capacity through conditional computation \citep{shazeer2017}. GShard extended sparse expert routing to large Transformer models and distributed training \citep{lepikhin2021gshard}, while Switch Transformers simplified routing to a single active expert and emphasized expert-capacity and load-balancing issues \citep{fedus2022}. Subsequent work such as ST-MoE further studied training stability and transfer behavior in sparse expert Transformers \citep{zoph2022}. Vision MoE extends sparse expert layers to image classification and studies performance--compute trade-offs in a non-language setting \citep{riquelme2021}. Expert-choice routing instead lets each expert select tokens, enforcing fixed expert loads while allowing a variable number of experts per token \citep{zhou2022expertchoice}. MegaBlocks addresses the systems trade-off between dropping overflow tokens and padding expert batches through block-sparse computation that does not drop tokens \citep{gale2023megablocks}. Our no-drop configurations use sufficient assignment capacity rather than the MegaBlocks implementation. These studies motivate our separation of stored expert capacity, active expert computation, and routing capacity when evaluating MoEParT.

\paragraph{Mixture-of-experts models in high-energy physics.}
Mixture-of-experts architectures have also begun to appear in high-energy physics. Genovese et al. combine graph Transformers with MoE layers for event-level signal-background classification, emphasizing expert specialization and interpretability \citep{genovese2025}. Subsequent work applies an MoE-enhanced graph Transformer to rare signatures in simulated ATLAS calorimeter data \citep{genovese2026calorimeter}. Our setting differs in that we study sparse expert routing within a Particle Transformer operating on constituent-level jet representations. To our knowledge, prior work has not systematically studied sparse MoE routing within the Particle Transformer architecture for large-scale jet classification.

\paragraph{Routing structure and expert specialization.}
Recent analysis of language-model MoEs relates expert usage to hidden-state geometry and cautions that routing patterns need not admit an interpretation in terms of domain expertise \citep{wang2026myth}. This motivates distinguishing routing association from semantic specialization. Our particle-level diagnostics quantify associations with identifiable physical properties; they do not establish the causal contribution or semantic role of individual experts.

\section{Model, data, and evaluation}
\paragraph{Architecture and routing.}
Both architectures embed 17 particle features with widths $(128,512,128)$ and use eight heads, eight particle-attention blocks, and two class-attention blocks. Pair embedding widths are $(64,64,64,8)$; the classifier has a 512-unit hidden layer and 188 outputs. Dense feed-forward networks and individual experts use $128\!\to\!512\!\to\!128$ with GELU. Particle and class-attention blocks use dropout 0.1 and 0, respectively. \moe\ replaces feed-forward networks in all ten blocks with $E$ experts.

\begin{figure*}[t]
    \centering
    \includegraphics[width=0.95\textwidth]{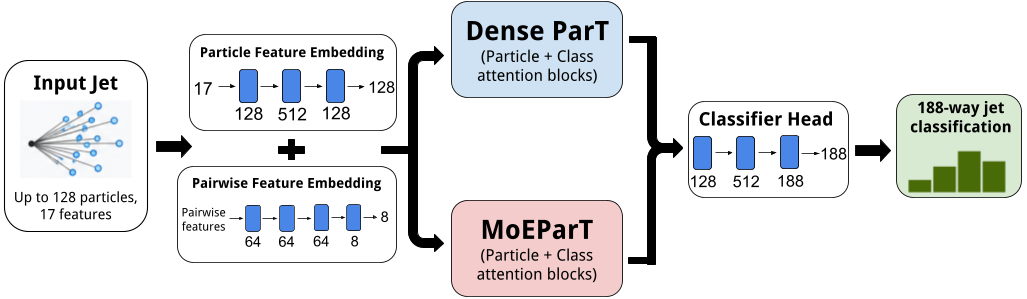}
    \caption{
    Overview of the dense Particle Transformer and MoEParT architectures. Dense ParT and MoEParT share the embedding and classifier architecture; MoEParT differs by replacing the dense feed-forward network with a sparsely routed mixture of experts. Both models produce a single 188-class jet prediction.
    }
    \label{fig:architecture}
\end{figure*}

For a token $x$, a learned linear router produces $p(x)=\operatorname{softmax}(Wx+b)$ and selects the $K$ largest probabilities, where $K$ is the number of active experts per token. For $T$ flattened tokens, the per-expert assignment capacity is
\begin{equation}
 C=\left\lfloor c\left\lceil TK/E\right\rceil\right\rfloor\qquad
 \label{eq:route}
\end{equation}
where \(E\) is the number of experts and \(c\) is the capacity factor; assignments exceeding an expert’s capacity are dropped. For $K>1$, the selected routing weights are renormalized before combining expert outputs, so increasing $K$ changes both active computation and how expert contributions are weighted. We also use an auxiliary load-balancing loss, weighted by \(\alpha\), to discourage highly imbalanced routing.

The load-balancing term for each particle-attention MoE block follows the Switch Transformer formulation \citep{fedus2022}, extended here to normalized top-$K$ assignment fractions:
\[
L_{\mathrm{bal}}
=E\sum_{e=1}^{E}\bar p_e f_e ,
\]
where $\bar p_e$ is the mean router probability for expert $e$ and $f_e$ is
the fraction of requested assignments to that expert before capacity filtering.
For $K>1$, assignment fractions are normalized by $K$. The training loss is
\[
L=L_{\mathrm{CE}}+
\alpha\sum_{\ell=1}^{8}L_{\mathrm{bal}}^{(\ell)},
\]
where the sum runs over the eight particle-attention blocks. Router jitter of
0.01 is applied only during training.

\begin{figure}[t]
    \centering
    \includegraphics[width=0.72\textwidth]{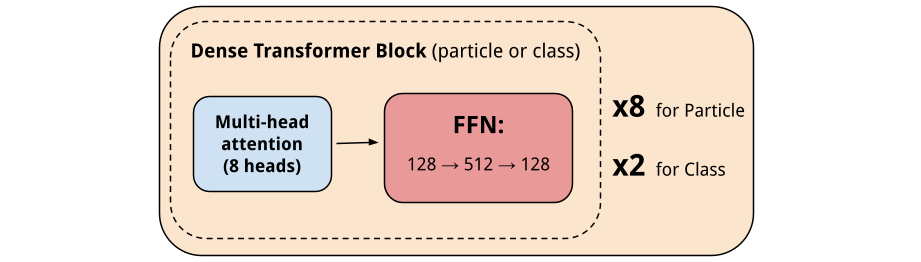}
    \hfill
    \includegraphics[width=0.82\textwidth]{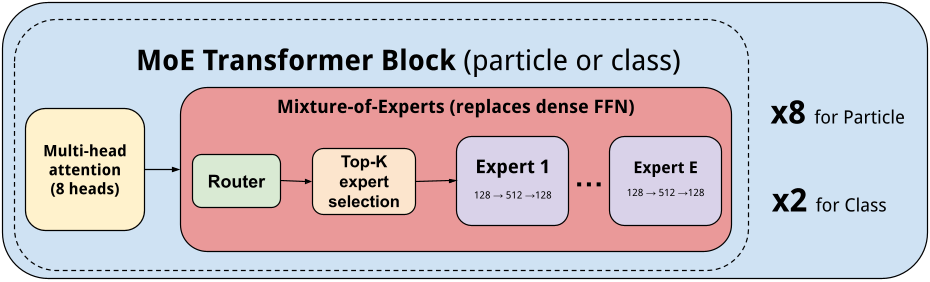}

    \caption{
    Dense and mixture-of-experts transformer blocks.
    Top: dense ParT uses a feed-forward network after attention. Bottom: MoEParT replaces this FFN with $E$ expert networks of the same architecture and routes each token to the top-$K$ selected experts. The replacement is applied in all eight particle-attention and two class-attention blocks.
    }
    \label{fig:block-comparison}
\end{figure}

\paragraph{Task and samples.}
We study 188-class jet classification on the Pythia JetClass-II dataset. The labels consist of two-prong resonances (0--14), three- and four-prong resonances (15--160), and QCD backgrounds (161--187). Each jet is represented by up to 128 particles using kinematic, particle-identity, charge, and displacement features, with preprocessing defined by the JetClass-II configuration.

Final predictive performance is evaluated on a held-out test subset containing 456,822 selected jets, of which 353,206 are signal and 103,616 are QCD. Checkpoints are selected using validation accuracy, while routing and representation analyses are performed on validation data to avoid using the held-out test set for model interpretation. All reported test metrics use the frozen selected checkpoint for each configuration.

\paragraph{Training setup.}
All dense and MoE configurations were trained using the same JetClass-II training and validation subsets and matched optimization budgets. The training subset consisted of 30 Parquet files: 10 each from the Res2P, Res34P, and QCD samples, corresponding to 3.0 million raw jets before the JetClass-II event selection. The validation subset consisted of six files, two from each sample group, corresponding to 600,000 raw jets before selection.

Models were trained for 20 epochs with 1,000,000 sampled training examples per epoch and 600,000 validation examples per epoch, using a batch size of 512. Optimization used Ranger with an initial learning rate of $5\times10^{-4}$. All model configurations used the same training files, validation files, samples per epoch, epoch count, batch size, and optimization settings, so comparisons between dense and MoE configurations use matched data exposure and training budgets. Checkpoints were selected according to validation accuracy. Each reported configuration corresponds to a single training run; variation across random seeds was not evaluated.

\paragraph{Experimental configurations.}
We compare dense ParT with four MoE studies: (i) capacity-limited top-1 models with \(E\in\{2,4,8\}\) and \(c=1.5\); (ii) no-drop top-1 models with \(c=E\); (iii) top-\(K\) models \((E,K,c)=(4,2,2),(8,2,4),(8,4,2)\); and (iv) an auxiliary-loss sweep at \((E,K,c)=(4,1,4)\) with \(\alpha\in\{0,0.01,0.05\}\). Unless stated otherwise, \(\alpha=0.01\); this checkpoint is shared with the no-drop E4/K1 model.

\paragraph{Evaluation metrics.}
We report 188-class classification accuracy together with signal-versus-QCD discrimination metrics. The signal score is defined as $s=\sum_{j=0}^{160} p_j,$ where classes \(0\!-\!160\) are treated as signal and classes \(161\!-\!187\) as QCD background. From this score we compute the binary AUROC and the QCD rejection $R_{\mathrm{QCD}}=\frac{1}{\epsilon_{\mathrm{QCD}}},$ evaluated at target signal efficiencies of 50\% and 80\%. Operating points are taken from the ROC point nearest the requested signal efficiency.

\paragraph{Computation and routing diagnostics.}
Table~\ref{tab:results} reports parameter counts and standardized forward-pass FLOPs measured for a single jet with 128 particles using \texttt{torch.profiler}. Fused scaled-dot-product attention is disabled only during profiling to expose the relevant operations. These FLOP counts measure arithmetic cost and should not be interpreted as wall-clock latency or throughput.

To quantify routing structure, we compute normalized mutual information
\[
\mathrm{NMI}(X,Y)=\frac{I(X;Y)}{\sqrt{H(X)H(Y)}} ,
\]
using the geometric-mean entropy normalization \citep{vinh2010} and natural-log entropies between the primary selected expert and particle-level properties such as particle type, relative \(p_T\), and \(p_T\) rank. Higher NMI indicates a stronger association between a particle property and the router's expert choice. We compare these aggregate associations across models rather than matching individual expert identities, since expert labels are arbitrary across independently trained runs. Padded particle tokens are excluded from the interpretability sample.

\section{Results}

\begin{figure*}[!htbp]
    \centering
    \includegraphics[width=\textwidth]{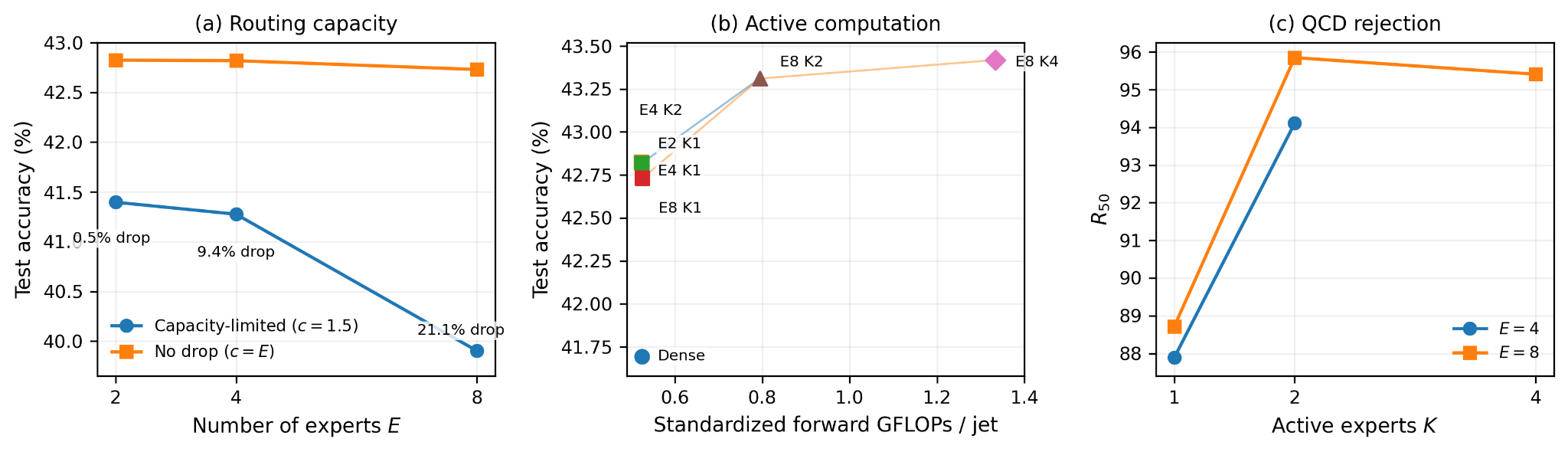}
    \caption{Performance under changes in routing capacity and active computation.
    (a) Capacity-limited and no-drop top-1 expert scaling; annotations give valid-assignment drop rates.
    (b) Test accuracy versus standardized forward GFLOPs.
    (c) \(R_{50}\) versus active experts.}
    \label{fig:performance}
\end{figure*}

\paragraph{Sparse conditional capacity improves over dense at nearly fixed compute.}
The no-drop top-1 MoE models outperform the dense baseline while using nearly identical standardized forward FLOPs (Fig.~\ref{fig:performance}(b)). Across \(E=2,4,8\), test accuracy is \(42.73\)–\(42.82\%\), compared with \(41.69\%\) for dense ParT, corresponding to an improvement of roughly one percentage point. Because only one expert is active per token, increasing \(E\) primarily increases stored parameter capacity rather than active computation. This suggests that sparse conditional capacity can improve predictive performance without requiring a comparable increase in nominal forward compute.

\begin{table}[t]
\centering
\small
\setlength{\tabcolsep}{4.2pt}
\begin{tabular}{lrrrrrrrrrr}
\toprule
Model & $E$ & $K$ & $c$ & $\alpha$ & $P$ & $G$ & Drop (\%) & Acc. (\%) & $R_{50}$ & $R_{80}$\\
\midrule
Dense & -- & -- & -- & -- & 2.30 & 0.524 & -- & 41.694 & 75.74 & 7.99\\
Limited & 2 & 1 & 1.5 & .01 & 3.64 & 0.524 & 0.465 & 41.396 & 74.76 & 8.00\\
Limited & 4 & 1 & 1.5 & .01 & 6.30 & 0.525 & 9.370 & 41.276 & 75.91 & 8.00\\
Limited & 8 & 1 & 1.5 & .01 & 11.62 & 0.526 & 21.126 & 39.902 & 54.53 & 6.31\\
\midrule
No drop & 2 & 1 & 2 & .01 & 3.64 & 0.524 & 0 & 42.824 & 83.03 & 8.28\\
No drop & 4 & 1 & 4 & .01 & 6.30 & 0.525 & 0 & 42.819 & 87.88 & 8.50\\
No drop & 8 & 1 & 8 & .01 & 11.62 & 0.526 & 0 & 42.730 & 88.71 & 8.45\\
Top-$K$ & 4 & 2 & 2 & .01 & 6.30 & 0.794 & 0 & 43.311 & 94.11 & 8.70\\
Top-$K$ & 8 & 2 & 4 & .01 & 11.62 & 0.795 & 0 & 43.311 & 95.85 & 8.91\\
Top-$K$ & 8 & 4 & 2 & .01 & 11.62 & 1.333 & 0 & 43.419 & 95.41 & 9.06\\
Auxiliary & 4 & 1 & 4 & 0 & 6.30 & 0.524 & 0 & 42.966 & 88.33 & 8.51\\
Auxiliary & 4 & 1 & 4 & .05 & 6.30 & 0.524 & 0 & 42.867 & 94.20 & 9.03\\
\bottomrule
\end{tabular}
\caption{Model configurations and held-out test performance.
$E$: number of experts; $K$: active experts per token;
$c$: capacity factor; $\alpha$: auxiliary-loss coefficient;
$P$: total parameters in millions; $G$: standardized forward
GFLOPs per jet at 128 particles; Drop: validation valid-assignment
drop rate. $R_{50}$ and $R_{80}$ denote QCD rejection at
50\% and 80\% signal efficiency, respectively.}
\label{tab:results}
\end{table}

\paragraph{Additional stored experts show diminishing accuracy gains.}
Within the no-drop \(K=1\) sweep, increasing \(E\) from 2 to 8 raises the parameter count from 3.64M to 11.62M while leaving active FLOPs essentially unchanged, yet accuracy remains nearly flat. Thus, the gain from sparse conditional capacity appears to saturate quickly in accuracy: E4 and E8 do not improve on E2 despite substantially greater stored capacity. The picture is more nuanced for background rejection, where \(R_{50}\) increases from 83.03 to 88.71, indicating that additional experts can still alter discrimination behavior even when top-line multiclass accuracy has saturated.

\paragraph{Routing capacity confounds expert-count scaling.}
The behavior of the expert-count sweep changes substantially when routing capacity is constrained. With a fixed capacity factor \(c=1.5\), the fraction of dropped valid assignments grows rapidly with expert count, reaching \(21.1\%\) for \(E=8\) (Fig.~\ref{fig:performance}(a)). Performance deteriorates in the same regime: the capacity-limited E8 model reaches \(39.90\%\) accuracy, compared with \(42.73\%\) when sufficient capacity eliminates dropping, while \(R_{50}\) increases from \(54.53\) to \(88.71\). Similar, though smaller, recoveries occur for E2 and E4. Thus, comparisons across expert count can be strongly confounded by routing capacity; increasing \(E\) without sufficient capacity simultaneously changes the available parameter capacity and the fraction of assignments that are successfully processed. Because the capacity-limited and no-drop models are separately trained, these results establish an association between overflow and degradation rather than a causal effect of token dropping alone.

\paragraph{More active experts improve performance at higher compute.}
Increasing \(K\) from 1 to 2 improves accuracy for both E4 and E8, but requires roughly 51\% more nominal forward FLOPs (Fig.~\ref{fig:performance}(b)). This indicates that the gains from top-2 routing come from a higher-compute regime rather than from additional stored capacity alone. Increasing E8 from \(K=2\) to \(K=4\) yields only a further \(0.11\) percentage-point accuracy improvement despite approximately 68\% more computation, suggesting diminishing returns from activating additional experts. The ranking also depends on the evaluation metric: E8/K4 slightly improves accuracy and \(R_{80}\), while E8/K2 retains the best \(R_{50}\) (Fig.~\ref{fig:performance}(c)). Thus, increasing active expert computation can improve predictive performance, but no single configuration dominates across all operating points.

\begin{figure*}[t]
    \centering
    \includegraphics[width=\textwidth]{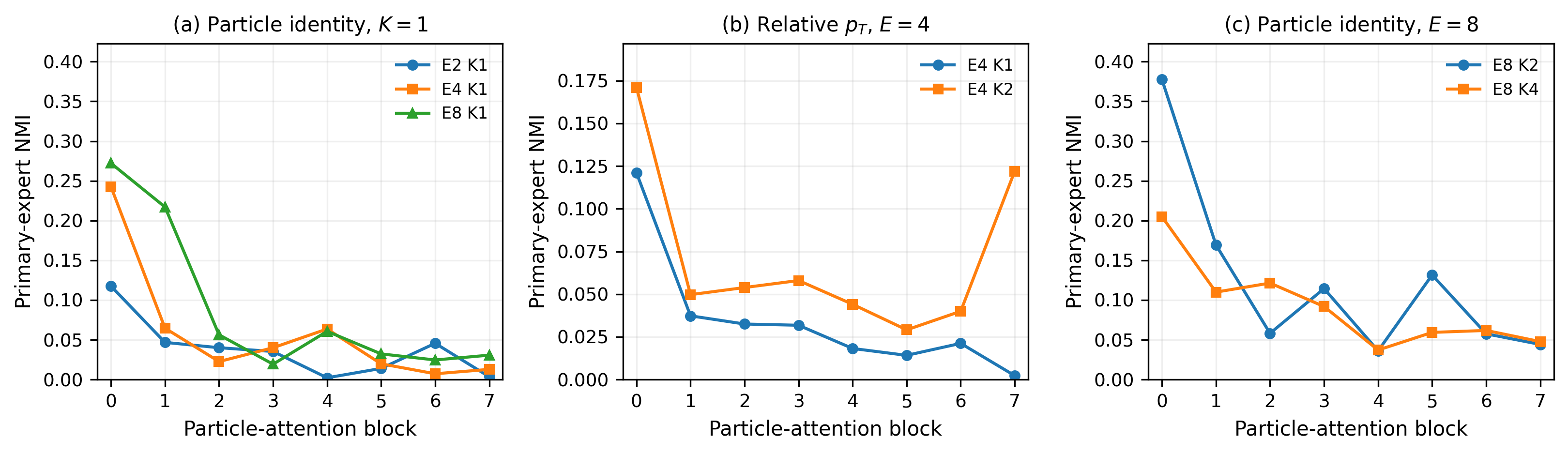}
    \caption{Layer-wise routing NMI on validation data.
    (a) Particle-type association for no-drop top-1 models.
    (b) Relative-\(p_T\) association for E4 K1/K2.
    (c) Particle-type association for E8 K2/K4.}
    \label{fig:routing}
\end{figure*}

\paragraph{Routing structure changes independently of predictive performance.}
Across the no-drop \(K=1\) sweep, accuracy is nearly unchanged while routing becomes more strongly associated with particle identity (Fig.~\ref{fig:routing}(a)); first-block particle-type NMI rises from \(0.118\) at \(E=2\) to \(0.272\) at \(E=8\). Thus, routing organization continues to change after accuracy has saturated, although these associations do not establish semantic expert specialization.

At \(E=4\), moving from \(K=1\) to \(K=2\) substantially strengthens late-layer relative-\(p_T\) routing (Fig.~\ref{fig:routing}(b)) and other kinematic and class-label associations. However, stronger routing association is not a monotonic predictor of performance: E8/K4 slightly improves accuracy over E8/K2 while first-block particle-type NMI falls from \(0.378\) to \(0.205\) (Fig.~\ref{fig:routing}(c)). Routing therefore characterizes how representations are partitioned across experts, but does not by itself explain predictive quality.

\paragraph{Auxiliary loss has metric-dependent effects.}
At fixed E4/K1, varying \(\alpha\) changes QCD rejection and routing association without a monotonic accuracy trend. In particular, \(\alpha=0.05\) improves rejection while producing lower first-block particle-type NMI, indicating that stronger load balancing does not consistently improve either accuracy or routing structure.

\section{Discussion and conclusion}

Our results show that mixture-of-experts Particle Transformers should not be characterized by expert count alone. Under top-1 routing, increasing the number of stored experts expands conditional parameter capacity while leaving active computation nearly unchanged, yet the corresponding predictive gains saturate quickly. In contrast, activating multiple experts per token improves performance while increasing active computation. These behaviors separate two distinct forms of MoE scaling: increasing the parameters available to the router and increasing the amount of expert computation applied to each token.

Routing capacity is also essential to this comparison. Increasing expert count without sufficient assignment capacity can cause a growing fraction of routed tokens to be dropped, confounding the effect of additional model capacity with reduced effective computation. Meaningful comparisons between sparse expert models therefore require considering not only how many experts are available, but also whether the routing mechanism can make use of them.

The routing analysis provides a complementary perspective. Expert assignments develop measurable associations with particle identity and kinematic properties, showing that the router learns structured partitions of particle representations. However, stronger routing structure does not monotonically correspond to higher classification performance. Routing organization is therefore informative about how the model allocates representations across experts, but should not be treated as a direct proxy for predictive quality or semantic expert specialization.

Overall, the effectiveness of MoEParT depends on the interaction between stored parameter capacity, active computation, and routing capacity. Sparse conditional computation can improve over a dense Particle Transformer at comparable nominal forward compute, but simply adding more experts is not sufficient; how those experts are activated and whether routed assignments are successfully processed are equally important.

\section{Limitations and Future Work}

This study focuses on a single Particle Transformer backbone and JetClass-II training setup, so the observed trade-offs may not transfer unchanged to other architectures or jet-classification tasks. In addition, the reported configurations are currently represented by a single trained checkpoint, leaving variation across random seeds unmeasured. The current study also uses a restricted training subset relative to the full JetClass-II training sample, which limits conclusions about how sparse expert capacity behaves in a higher-data regime. Future work will therefore evaluate the principal dense and MoE configurations across multiple seeds and larger training subsets to determine the stability of the observed performance and routing trends.

Beyond addressing these scope limitations, several extensions could help clarify when and why sparse expert routing is useful. One direction is to study data scaling explicitly, comparing dense and MoE models as the available training sample increases to determine whether additional expert capacity becomes more valuable when data is less limiting. Another is to investigate routing dynamics during training rather than only at the selected checkpoint, which could reveal when routing structure emerges and whether it stabilizes, collapses, or reorganizes over time. Since NMI is normalized but not adjusted for chance \citep{vinh2010}, future comparisons across expert counts could include shuffled-label baselines or adjusted mutual information. Functional interventions would also help distinguish routing associations from expert specialization.

\paragraph{Code and reproducibility.}
Code, model configurations, and training scripts used in this study are
available at \url{https://github.com/kpendiyala/MPT}. The corresponding repository revision is \texttt{df27f8e912a08ea887f77c7940e3932e2a47fc11}. The repository includes the dense and MoE Particle Transformer implementations, JetClass-II training configuration, and experiment settings used for the reported results.

\clearpage
\bibliographystyle{plainnat}
\bibliography{refs}\end{document}